\documentclass{article}
\usepackage[final]{nips_2017}
\usepackage{natbib}
\setcitestyle{round}

\usepackage[utf8]{inputenc} 
\usepackage[T1]{fontenc}    

\usepackage{hyperref}       
\usepackage{url}            
\usepackage{booktabs}       
\usepackage{amsfonts}       
\usepackage{nicefrac}       
\usepackage{microtype}      

\usepackage{isomath}
\usepackage{amsfonts}
\usepackage{amssymb}
\usepackage{amsmath}
\usepackage[pdftex]{graphicx}
\usepackage{algorithm}
\usepackage[noend]{algpseudocode}
\usepackage{multirow}

\title{ShiftCNN: Generalized Low-Precision Architecture for Inference of Convolutional Neural Networks}

\author{Denis~A.~Gudovskiy,~Luca~Rigazio\\
Panasonic Silicon Valley Lab\\
Cupertino, CA, 95014\\
\texttt{\{denis.gudovskiy,luca.rigazio\}@us.panasonic.com} \\
}

\newcommand{\mytensor}[1]{\boldsymbol{\mathsf{#1}}}

\newcommand{\myvector}[1]{\boldsymbol{\mathit{#1}}}

\newcommand{\specialcell}[2][c]{%
	\begin{tabular}[#1]{@{}c@{}}#2\end{tabular}}

\begin{document}

\maketitle

\begin{abstract}
In this paper we introduce ShiftCNN, a generalized low-precision architecture for inference of multiplierless convolutional neural networks (CNNs). ShiftCNN is based on a power-of-two weight representation and, as a result, performs only shift and addition operations. Furthermore, ShiftCNN substantially reduces computational cost of convolutional layers by precomputing convolution terms. Such an optimization can be applied to any CNN architecture with a relatively small codebook of weights and allows to decrease the number of product operations by at least two orders of magnitude. The proposed architecture targets custom inference accelerators and can be realized on FPGAs or ASICs. Extensive evaluation on ImageNet shows that the state-of-the-art CNNs can be converted without retraining into ShiftCNN with less than 1\% drop in accuracy when the proposed quantization algorithm is employed. RTL simulations, targeting modern FPGAs, show that power consumption of convolutional layers is reduced by a factor of 4 compared to conventional 8-bit fixed-point architectures.
\end{abstract}

\section{Introduction}
\label{sec:intro}
Current achievements of deep learning algorithms made them an attractive choice in many research areas including computer vision applications such as image classification~\citep{he}, object detection~\citep{yolo}, semantic segmentation~\citep{long} and many others. Until recently real-time inference of deep neural networks (DNNs) and, more specifically, convolutional neural networks (CNNs) was only possible on power-hungry graphics processing units (GPUs). Today numerous research and implementation efforts~\citep{survey} are conducted in the area of computationally inexpensive and power-efficient CNN architectures and their implementation for embedded systems as well as data center deployments.

We propose a generalized low-precision CNN architecture called ShiftCNN. ShiftCNN substantially decreases computational cost of convolutional layers which are the most computationally demanding component during inference phase. The proposed architecture combines several unique features. First, it employs a hardware-efficient power-of-two weight representation which requires performing only shift and addition operations. State-of-the-art CNNs can be converted into ShiftCNN-type of model without network retraining with only a small drop in accuracy. Second, a method to precompute convolution terms decreases the number of computations by at least two orders of magnitude for commonly used CNNs. Third, we present ShiftCNN design which achieves a factor of 4 reduction in power consumption when implemented on a FPGA compared to conventional 8-bit fixed-point architectures. In addition, this architecture can be extended to support sparse and compressed models according to conventional~\citep{obd} or more recent approaches~\citep{dcm}.

The rest of the paper is organized as follows. Section~\ref{sec:related} reviews related work. Section~\ref{sec:optimality} gives theoretical background. Section~\ref{sec:quantization} proposes a generalized low-precision quantization algorithm. Section~\ref{sec:architecture} introduces a method to decrease computational cost of convolutional layers and its corresponding architecture. Section~\ref{sec:evaluation} presents experimental results on ImageNet, quantization and complexity analysis including power consumption projections for FPGA implementation. The last section draws conclusions.

\section{Related Work}
\label{sec:related}
\subsection{Low-Precision Data Representation}
Conventional CNN implementations on GPUs use a power-inefficient floating-point format. For example, \citep{horowitz} estimated that 32-bit floating-point multiplier consumes 19$\times$ more power than 8-bit integer multiplier. Several approaches for a more computationally-efficient data representation within CNNs have been proposed recently. One direction is to convert (quantize) a baseline floating-point model into a low-precision representation using quantization algorithms. Such quantization algorithms can be classified into the following categories: fixed-point quantization~\citep{CourbariauxBD14}, binary quantization~\citep{courbariauxB16,RastegariORF16}, ternary quantization~\citep{liL16} and power-of-two quantization~\citep{MiyashitaLM16,zhou2017}. Moreover, these quantization algorithms can be selectively applied to various parts of the network e.g. weights, inputs and feature maps.

According to the latest published experimental results~\citep{zhou2017}, out of the referenced quantization algorithms only models converted to fixed-point and power-of-two representations achieve baseline accuracy without significant accuracy drop. In addition, binary and ternary-quantized models require a subset of convolutional layers, e.g. first and last layers, to be in the other, usually fixed-point, format. In this paper, we select power-of-two weight representation for building truly multiplierless CNNs and, unlike~\citep{MiyashitaLM16,zhou2017}, employ a more general quantization algorithm which allows to achieve minimal accuracy drop without time-consuming retraining. In our architecture, inputs and feature maps are in the dynamic fixed-point format.

\subsection{Convolution Computation Algorithms}
There is a significant body of work on various convolution computation algorithms e.g. frequency-domain convolution or Winograd minimal filtering convolution summarized in~\citep{Lavin15b}. Unfortunately, they cannot be universally applied for all CNN architectures. For example, frequency-domain methods are not suitable for modern CNNs with small convolution filter sizes and Winograd algorithm achieves moderate speed-ups (6-7$\times$) only with batch sizes more than one which increases processing latency. Other algorithms suited for GPUs with optimized vector operations using BLAS libraries are not compatible with custom ASIC or FPGA accelerators.

Another interesting approach was proposed by~\citep{lcnn} where convolutions are optimized by lookups into a shared trained dictionary assuming that weights can be represented by a trained linear combinations of that dictionary. While this approach achieves substantial speed-ups during inference, it significantly degrades accuracy of networks due to inherent weight approximation assumptions. Instead, our approach employs powerful distribution-induced representation to form a naturally small codebook (dictionary) . 

\section{Note on Quantization Optimality}
\label{sec:optimality}
Consider a general feedforward non-linear neural network that maps an input $\myvector{x}$ to an output $\myvector{y}$ according to
\begin{equation} \label{eq:01}
\myvector{y} = f(\myvector{x};\boldsymbol{\theta}),
\end{equation}
where $\boldsymbol{\theta}$ is the vector of all trainable parameters and $f$ is the function defined by the network structure. Typically, the parameters $\boldsymbol{\theta}$ are found using maximum log-likelihood criteria with regularization by solving optimization problem defined by
\begin{equation} \label{eq:02}
\arg \max_{\boldsymbol{\theta}}  \log p(\myvector{y} | \myvector{x};\boldsymbol{\theta}) - \alpha R(\boldsymbol{\theta}),
\end{equation}
where $R(\boldsymbol{\theta})$ is a regularization term that penalizes parameters and $\alpha \geq 0 $ is a hyperparameter that defines a tradeoff between estimator bias and variance. Usually, only a subset of $\boldsymbol{\theta}$ i.e. the weights $\myvector{w}$ of the model are penalized. Hence, the most widely used $L_2$ and $L_1$ regularization types define regularization term as $R(\boldsymbol{\theta})= \| \myvector{w} \|_2^2$ and $R(\boldsymbol{\theta})= \| \myvector{w} \|_1$, respectively. It is well-known~\citep{goodbook} that regularization places a prior on weights which induces normal distribution for $L_2$ regularization and Laplace distribution for $L_1$ regularization.

Based on the known probability density function (PDF) $f_{W}(w)$, we can estimate quantization distortion by
\begin{equation} \label{eq:03}
D = \sum_{j=1}^{J} \int_{w_{j-1}}^{w_{j}} |w-\tilde{w}_j|^2 f_{W}(w) dw,
\end{equation}
where $J$ is the size of quantizer codebook. Then, the optimal quantizer in the mean squared error (MSE) sense minimizes (\ref{eq:03}). By taking into account (\ref{eq:02}), the distortion measure $D$ is minimized when a non-uniform quantizer is used presuming selected regularization method. Interestingly, a similar reasoning can be applied to network feature maps. Although, we derived the optimal quantizer in the MSE sense, some network parameters may have more effect on the optimization objective function (\ref{eq:02}) than others. For example,~\citet{limits} proposed to use a Hessian-weighted distortion measure in order to minimize the loss due to quantization assuming a second-order Taylor series expansion. Hence, an additional objective function-dependent term can be added to (\ref{eq:03}) as well.


\section{ShiftCNN Quantization}
\label{sec:quantization}
The input of each convolutional layer in commonly used CNNs can be represented by an input tensor $\mytensor{X}\in\mathbb{R}^{C \times H \times W}$, where $C$, $H$ and $W$ are the number of input channels, the height and the width, respectively. The input $\mytensor{X}$ is convolved with a weight tensor $\mytensor{W}\in\mathbb{R}^{\tilde{C} \times C \times H_f \times W_f}$, where $\tilde{C}$ is the number of output channels, $H_f$ and $W_f$ are the height and the width of filter kernel, respectively. A bias vector $\myvector{b}\in\mathbb{R}^{\tilde{C}}$ is added to the result of convolution operation. Then, the $\tilde{c}$th channel of an output tensor $\mytensor{Y}\in\mathbb{R}^{\tilde{C} \times \tilde{H} \times \tilde{W}}$ before applying non-linear function can be computed by
\begin{equation} \label{eq:1}
\mytensor{Y}_{\tilde{c}} = \mytensor{W}_{\tilde{c}} \ast \mytensor{X} + \myvector{b}_{\tilde{c}},
\end{equation}
where $\ast$ denotes convolution operation.

Based on conclusions in Section~\ref{sec:optimality}, we approximate $\mytensor{W}$ by a low-precision weight tensor $\hat{\mytensor{W}}$ using hardware-efficient non-uniform quantization method. Each entry $\hat{w}_i~(i\in\{\tilde{c},c,h_f,w_f\})$ of $\hat{\mytensor{W}}$ is defined by
\begin{equation} \label{eq:2}
\hat{w}_i = \sum_{n=1}^{N} \mathcal{C}_n [ \mathrm{idx}_i(n) ],
\end{equation}
where for ShiftCNN codebook set $\mathcal{C}_n=\{0,\pm 2^{-n+1},\pm 2^{-n},\pm 2^{-n-1},\ldots,\pm 2^{-n-\lfloor M/2 \rfloor+2}\}$ and a codebook entry can be indexed by $B$-bit value with total $M=2^B-1$ combinations. Then, each weight entry can be indexed by $(N B)$-bit value. For example, consider a case with $N=2$ and $B=4$. Then, the weight tensor is indexed by 8-bit value with codebooks $\mathcal{C}_1=\{0,\pm 2^{0},\pm 2^{-1},\ldots,\pm 2^{-6}\}$ and $\mathcal{C}_2=\{0,\pm 2^{-1},\pm 2^{-2},\ldots,\pm 2^{-7}\}$. Note that for $N=1$, (\ref{eq:2}) can be simplified to binary-quantized ($B=1$, $0 \notin \mathcal{C}_1$) model or ternary-quantized ($B=2$) model. Moreover, fixed-point integers can be represented by (\ref{eq:2}) as well when $N \gg 1$. Similar to~\citep{zhou2017}, we normalize $\mytensor{W}$ by $\max{( \mathrm{abs}(\mytensor{W}) )}$, where $\mathrm{abs}(\cdot)$ is a element-wise operation, before applying (\ref{eq:2}). Then, $N\times1$ weight index vector can be calculated by Algorithm~\ref{alg:1}. Unlike the quantization algorithms in~\citep{MiyashitaLM16,zhou2017} for $N=1$, the proposed algorithm is more general and suitable for cases $N>1$.

\begin{algorithm}
\caption{Quantization to ShiftCNN weight representation}
\label{alg:1}
\begin{algorithmic}[1]
\State \textbf{Initialize:} $\tilde{w}_i = 0$, $r = w_i / \max{( \mathrm{abs}(\mytensor{W}) )}$, where $i\in\{\tilde{c},c,h_f,w_f\}$
\For{$n=1 \ldots N$}
	\State $q_{\mathrm{sgn}} = \mathrm{sgn}(r)$
    \State $q_{\mathrm{log}} = \log_2 \lvert r \rvert$
    \State $q_{\mathrm{idx}} = \lfloor q_{\mathrm{log}} \rfloor$
    \State $b_{\mathrm{log}} = q_{\mathrm{idx}} + \log_2 1.5$
    \If{$q_{\mathrm{log}}>b_{\mathrm{log}}$}
    	\State $q_{\mathrm{idx}}\texttt{++}$
    \EndIf
    \State $q = q_{\mathrm{sgn}} 2^{q_{\mathrm{idx}}}$
    \State $\mathrm{idx}_i(n) = q_{\mathrm{sgn}} (-n-q_{\mathrm{idx}}+2)$ 
    \If{$\lvert \mathrm{idx}_i(n) \rvert > \lfloor M/2 \rfloor$}
    	\State $q = 0$
        \State $\mathrm{idx}_i(n) = 0$
    \EndIf
    \State $r~\texttt{-=}~q$
    \State $\tilde{w}_i~\texttt{+=}~q$
\EndFor
\end{algorithmic}
\end{algorithm}

\begin{figure}[ht]
	\begin{center}
		\includegraphics[width=0.7\linewidth]{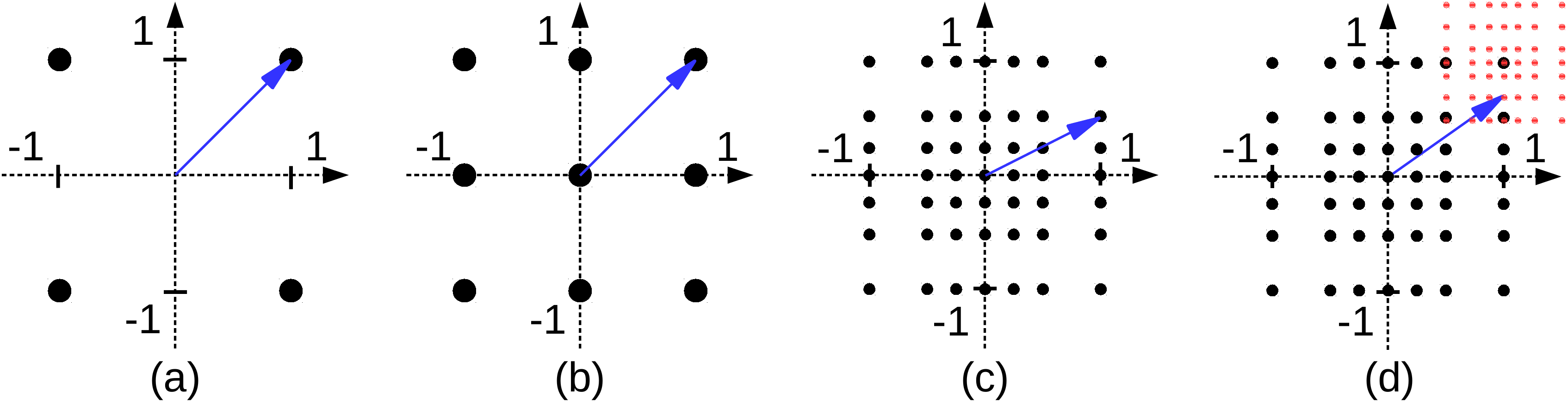}
		\caption{Weight vector in 2D space for models: (a) binary-quantized ($N=1,B=1$, $0 \notin \mathcal{C}_1$), (b) ternary-quantized ($N=1,B=2$), (c) ($N=1,B=3$) and (d) ($N=2,B=3$).}
		\label{fig:0}
	\end{center}
\end{figure}

Effectively, the non-uniform quantizer in Algorithm~\ref{alg:1} maps the weight tensor $\mytensor{W}\in\mathbb{R}^{\tilde{C} \times C \times H_f \times W_f} \to \hat{\mytensor{W}}\in\mathcal{C}_{N}^{\tilde{C} \times C \times H_f \times W_f}$, where $\mathcal{C}_{N}$ is a set formed by sum of $\mathcal{C}_{n}$ sets. Figure~\ref{fig:0} illustrates possible $\mathcal{C}_{N}$ sets i.e. weight fields. Interestingly, binary-quantized filters contribute the same unit magnitude to the output feature map with either positive or negative sign, while ternary-quantized weights allow to mute that contribution in some dimensions of weight vector space. In other words, output feature map heavily depend on input magnitudes. At the same time, ShiftCNN is more flexible and allows to trade off computational complexity and cardinality of weight field $\mathcal{C}_{N}$ while reusing the same arithmetic pipeline.

In this work, we mostly concentrate on a weight approximation task for hardware-efficient $\mathcal{C}_{N}$, while~\citep{courbariauxB16,RastegariORF16,liL16,zhou2017} empirically employ a small subset of $\mathcal{C}_{N}$ and, in order to restore desired accuracy, rely on a constrained retraining process i.e. introduce additional regularization. Though the topic of a high-dimensional classifier over reduced field needs further research, we can derive simple dimension relationships between $\mathcal{C}_{N}$ subfields using field extensions. For example, vector $\myvector{w} \in \mathcal{C}_{N}^V$ dimension can be rewritten over a ternary field $\mathcal{T}=\{0,\pm 1\}$ as
\begin{equation} \label{eq:3}
\dim_{\mathcal{C}_{N}}(\myvector{w}) = 
\underbrace{\dim_{\mathcal{C}_{N}}(\mathcal{T})}_\text{$\lfloor M/2 \rfloor+N-1$}
\underbrace{\dim_{\mathcal{T}}(\myvector{w})}_\text{$V$}.
\end{equation}

\section{ShiftCNN Architecture} 
\label{sec:architecture}

\begin{figure}[ht]
	\begin{center}
		\includegraphics[width=0.85\linewidth]{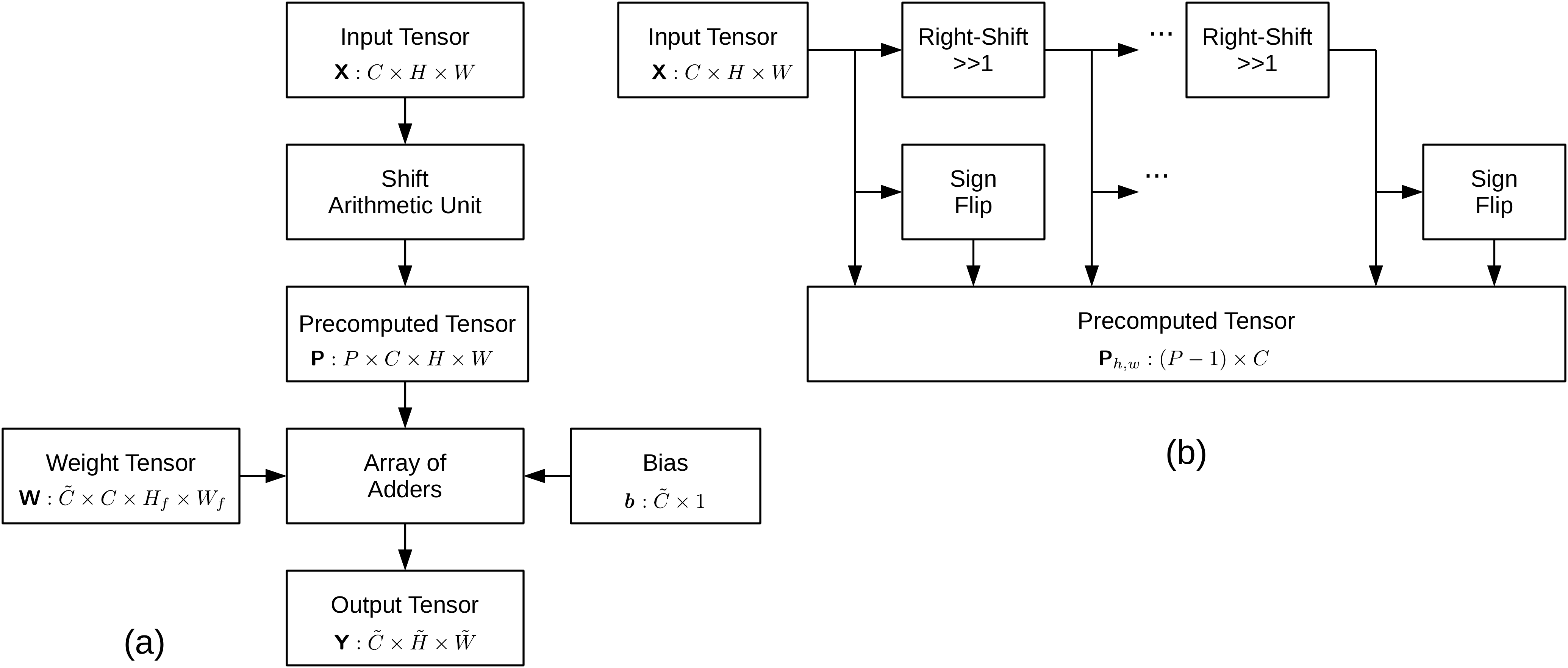}
		\caption{(a) High-level structure. (b) Shift Arithmetic Unit (ShiftALU).}
		\label{fig:1}
	\end{center}
\end{figure}

We propose to precompute all possible $P=(M+2(N-1))$ combinations of product terms for each element in the input tensor $\mytensor{X}$ when (\ref{eq:1}) is applied and, then, accumulate results in the output tensor $\mytensor{Y}$ according to the weight tensor $\hat{\mytensor{W}}$ indices. Assuming for simplicity $\tilde{H}=H$, $\tilde{W}=W$ for filter kernels with stride $1$, conventional convolutional layer performs $ ( \tilde{C} H_f W_f C H W )$ multiplications. In the proposed approach, a convolutional layer computes only $ ( P C H W ) $ product operations, which for ShiftCNN are power-of-two shifts. For the state-of-the-art CNNs $ P \ll ( \tilde{C} H_f W_f ) $ and, therefore, the computational cost reduction in terms of the number of product operations is quantified by a $ ( \tilde{C} H_f W_f / P) $ ratio. Note that the number of addition operations is $N$-times higher for the proposed approach. In addition, a new memory buffer for precomputed tensor $\mytensor{P}\in\mathbb{R}^{ P \times C \times H \times W}$ is implicitly introduced. Details on further microarchitecture-level optimizations including the size of the $\mytensor{P}$ are given below.

\begin{figure}[ht]
	\begin{center}
		\includegraphics[width=0.7\linewidth]{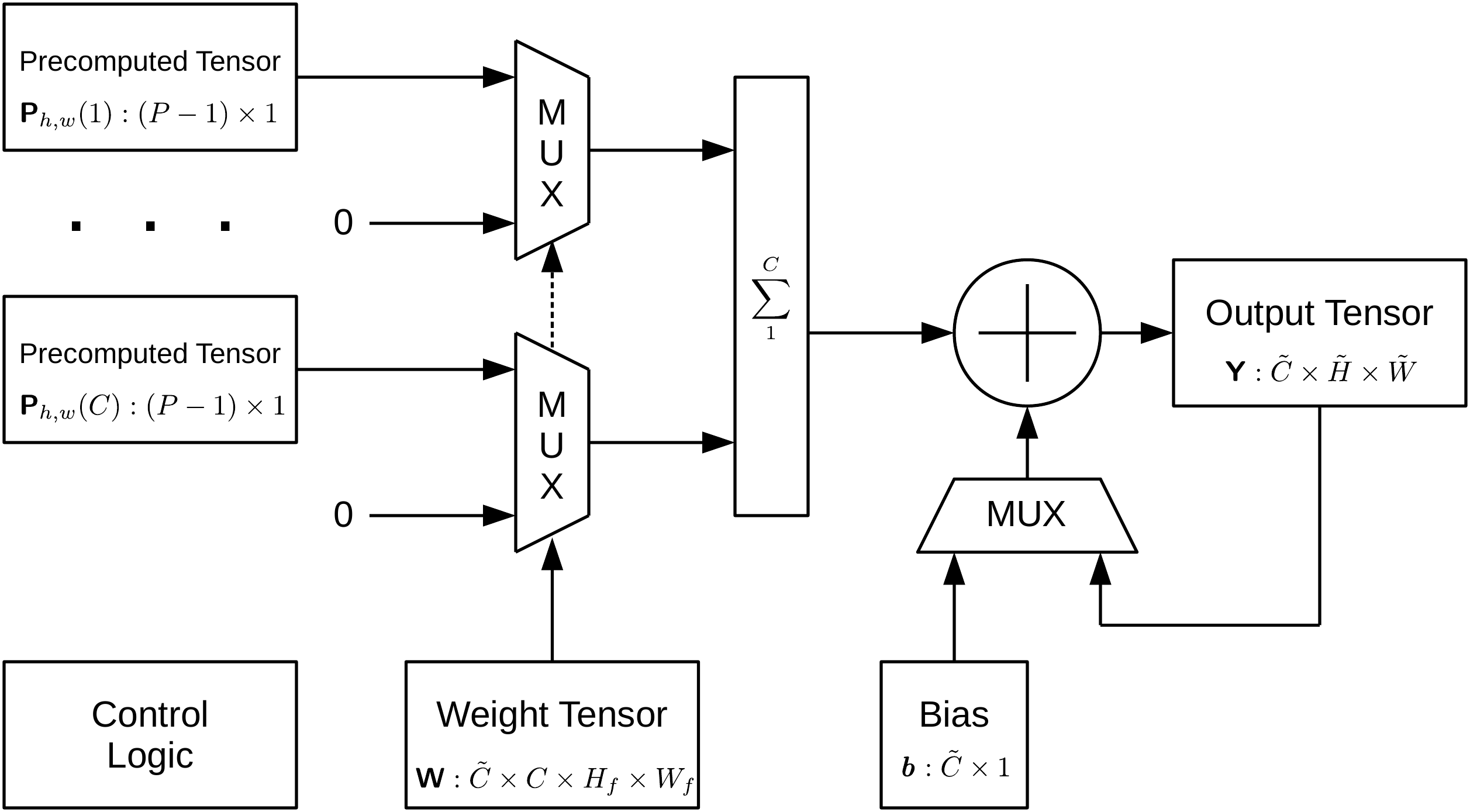}
		\caption{Array of adders.}
		\label{fig:2}
	\end{center}
\end{figure}

Figure~\ref{fig:1}(a) illustrates a high-level structure of the proposed ShiftCNN inference architecture. It computes (\ref{eq:1}) according to assumptions in (\ref{eq:2}) by precomputing tensor $\mytensor{P}$ on a special shift arithmetic unit (ShiftALU). The ShiftALU block showed in Figure~\ref{fig:1}(b) consists of a shifter and a sign flipping logic. The shifter shifts input values to the right and writes each shift-by-one result as well as pass-through input value to the memory buffer to keep precomputed $\mytensor{P}$ tensor. In addition, each of $\lfloor P/2 \rfloor$ precomputed terms is passed through the sign flipping logic and, then, written into $\mytensor{P}$ memory buffer. Mathematically, ShiftALU computations are equivalent to replacing the weight tensor $\mytensor{W}$ in (\ref{eq:1}) convolution with the codebook $\mathcal{C}_N$ in order to generate precomputed convolution terms.

\begin{algorithm}
	\caption{Scheduling and control logic}
	\label{alg:2}
	\begin{algorithmic}[1]
		\For{$h=1 \ldots H$ and $w = 1 \ldots W$}
		\State read $C \times 1$ input vector $\myvector{x}=\mytensor{X}_{h,w}$
		\State precompute $(P-1) \times C$ matrix $\myvector{P}=\mytensor{P}_{h,w} $ and write into the memory buffer
		\For{$\tilde{c}=1 \ldots \tilde{C}$}
		\State write bias $\myvector{b}_{\tilde{c}}$ to the output tensor $\mytensor{Y}_{\tilde{c},h,w}$
		\For{$ n=1 \ldots N $}
		\For{$h_f=1 \ldots H_f$ and $w_f=1 \ldots W_f$}
		\State select value from $\{0,\myvector{P}\}$ using $\mathrm{idx}_i(n)$, where $i\in\{\tilde{c},c,h_f,w_f\}$
		\State add value to the output tensor $\mytensor{Y}_{\tilde{c},h,w}$
		\EndFor
		\EndFor
		\EndFor
		\EndFor
	\end{algorithmic}
\end{algorithm}

Then, precomputed convolution terms are fetched from the memory buffer according to the weight tensor indices using a multiplexer as showed in Figure~\ref{fig:2} and processed on an array of adders. The multiplexer selects either one of $(P-1)$ memory outputs or zero value using $B$-bit index. Next, the multiplexer outputs are accumulated and bias term added according to scheduling and control logic. Lastly, the result is stored in the output tensor. One of possible scheduling and control logic schemes for ShiftCNN is presented in Algorithm~\ref{alg:2} pseudo-code. It extracts computational parallelism along $C$-dimension of the input tensor. We define a parallelization level $\bar{C}$ as the number of concurrently computed convolution terms. For the sake of simplicity, we consider a case with stride 1 filter kernel and, therefore, $\tilde{H}=H$, $\tilde{W}=W$ and choose the parallelization level $\bar{C}=C$. Therefore, we are able to decrease size of the precomputed memory to $ ( (P-1) C ) $ entries. In order to provide enough bandwidth for the selected parallelization level, the memory buffer can be organized as an array of $(P-1)$ length-$C$ shift registers which are written along $C$ dimension and read (multiplexed) along $P$-dimension. Then, the array of adders for the parallelized case is naturally transformed into an adder tree. 


We can consider several corner cases for ShiftCNN architecture. For a case $N=1$ and $B=1$, the proposed shift arithmetic unit can be simplified into architecture for processing binary-quantized models with a single sign flipping block only. For a case $N=1$ and $B=2$, ShiftALU is the same as in the previous case, but multiplexer is extended by zero selection logic for processing ternary-quantized models. For a case $N \gg 1$, ShiftCNN can be converted into architecture for processing fixed-point models.

\section{Evaluation}
\label{sec:evaluation}

\subsection{ImageNet Results}

To evaluate the accuracy of ShiftCNN and its baseline full-precision counterparts, we selected the most popular up-to-date CNN models for image classification. Caffe~\citep{caffe} was used to run networks and calculate top-1 and top-5 accuracies. GoogleNet and ResNet-50 pretrained models were downloaded from the Caffe model zoo\footnote{https://github.com/BVLC/caffe/wiki/Model-Zoo}. SqueezeNet v1.1\footnote{https://github.com/DeepScale/SqueezeNet} and ResNet-18\footnote{https://github.com/HolmesShuan/ResNet-18-Caffemodel-on-ImageNet} were downloaded from the publicly available sources. All the referenced models were trained on ImageNet~\citep{imagenet} (ILSVRC2012) dataset by their authors. A script to convert baseline floating-point model into ShiftCNN without retraining is publicly available\footnote{https://github.com/gudovskiy/ShiftCNN}. The script emulates power-of-two weight representation by decreasing precision of floating-point weights according to Algorithm~\ref{alg:1}. We run 50,000 images from ImageNet validation dataset on the selected network models and report accuracy results in Table~\ref{tab:1}. ResNet results in parentheses include quantized batch normalization and scaling layers with another weight distribution as explained below.

\begin{table}[ht]
	\caption{ImageNet accuracy of baseline models and corresponding ShiftCNN variants.}
	\label{tab:1}
	\centering
	\begin{tabular}{lcccccc}
        \toprule
		Network & \specialcell{Shifts\\$N$} & \specialcell{Bit-\\width $B$} & \specialcell{Top-1\\Accuracy, \%} & \specialcell{Top-5\\Accuracy, \%} & \specialcell{Decrease\\in Top-1\\Accuracy, \%} & \specialcell{Decrease\\in Top-5\\Accuracy, \%} \\
		\midrule
        SqueezeNet & base & 32 & 58.4 & 81.02 & & \\
        SqueezeNet & 1 & 4 & 23.01 & 45.93 & 35.39 & 35.09 \\
        SqueezeNet & 2 & 4 & 57.39 & 80.31 & 1.01 & 0.71 \\
        SqueezeNet & 3 & 4 & 58.39 & 81.01 & 0.01 & 0.01 \\
        \midrule
        GoogleNet & base & 32 & 68.93 & 89.15 & & \\
        GoogleNet & 1 & 4 & 57.67 & 81.79 & 11.26 & 7.36 \\
        GoogleNet & 2 & 4 & 68.54 & 88.86 & 0.39 & 0.29 \\
        GoogleNet & 3 & 4 & 68.88 & 89.06 & 0.05 & 0.09 \\
        \midrule
        ResNet-18 & base & 32 & 64.78 & 86.13 & & \\
        ResNet-18 & 1 & 4 & 24.61 & 47.02 & 40.17 & 39.11 \\
        ResNet-18 & 2 & 4 & 64.24(61.57) & 85.79(84.08) & 0.54(3.21) & 0.34(2.05) \\
        ResNet-18 & 3 & 4 & 64.75(64.69) & 86.01(85.97) & 0.03(0.09) & 0.12(0.16) \\
        \midrule
        ResNet-50 & base & 32 & 72.87 & 91.12 & & \\
        ResNet-50 & 1 & 4 & 54.38 & 78.57 & 18.49 & 12.55 \\
        ResNet-50 & 2 & 4 & 72.20(70.38) & 90.71(89.48) & 0.67(2.49) & 0.41(1.64) \\
        ResNet-50 & 3 & 4 & 72.58(72.56) & 90.97(90.96) & 0.29(0.31) & 0.15(0.16) \\
		\bottomrule
	\end{tabular}
\end{table}

The converted models in Table~\ref{tab:1} have a performance drop of less than 0.29\% for convolutional layers when $B=4$ and $N>2$. The models with $B=4$ and $N=2$ have a performance drop within 1\% which is a good practical trade-off between accuracy and computational complexity. We didn't notice any significant accuracy improvements for bit-widths $B>4$ when $N>1$. While in Table~\ref{tab:1} the cases with $N=1$ do not perform well without retraining, \citet{zhou2017} reported that the cases with $N=1$ and $B=5$ perform as good as the baseline models with their retraining algorithm.

\subsection{Weight Quantization Analysis}

\begin{figure}[ht]
	\begin{center}
		\includegraphics[width=1.0\linewidth]{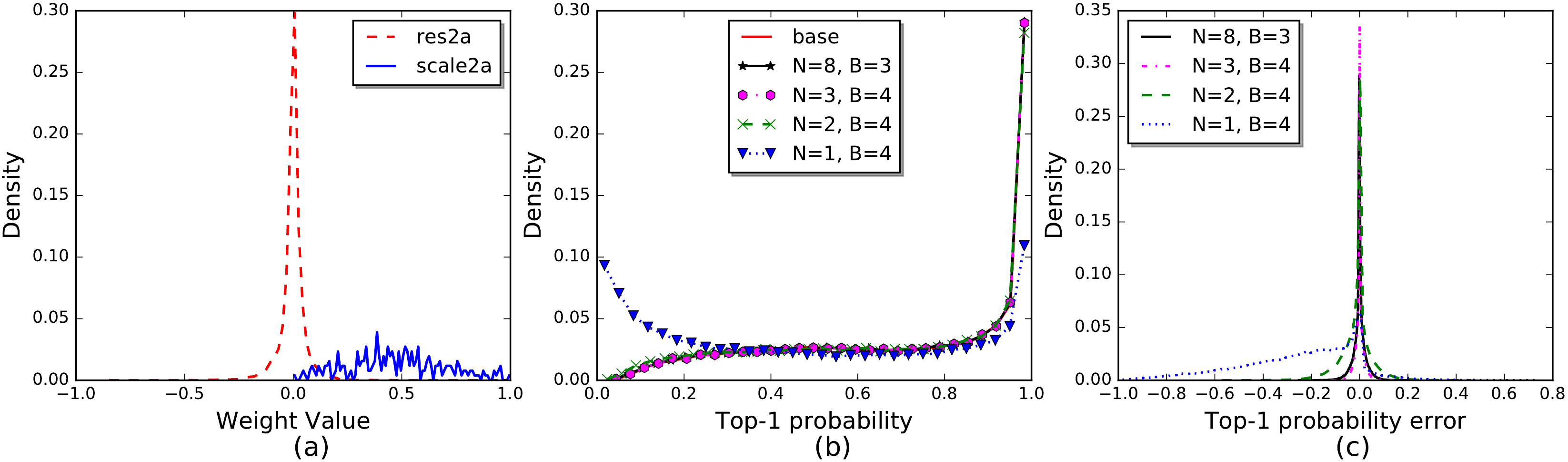}
		\caption{PDFs: (a) ResNet-50 weights, (b) GoogleNet top-1 class and (c) GoogleNet top-1 class error.}
		\label{fig:3}
	\end{center}
\end{figure}

Surprisingly good results for such low-precision logarithmic-domain representation even without retraining can be explained by looking at the distribution of the network weights. Figure~\ref{fig:3}(a) illustrates weight (normalized to unit magnitude) PDF of a typical convolutional layer (\textit{res2a\_branch1} layer in ResNet-50) as well as a scaling layer \textit{scale2a\_branch1} in ResNet-50. The typical convolutional layer weights are distributed according to Section~\ref{sec:optimality} which is well approximated by the non-uniform quantization Algorithm~\ref{alg:1}. At the same time, batch normalization and scaling layers in ResNet networks have an asymmetric and a broader distributions which cannot be approximated well enough unless $N>2$.

Due to difficulty of analytical analysis of chained non-linear layers, we empirically evaluate quantization effects at the GoogleNet softmax layer output with ImageNet validation dataset input. Figure~\ref{fig:3}(b) shows top-1 class PDF for baseline floating-point, 8-bit fixed-point ($N=8, B=3$) and other ShiftCNN configurations. According to that PDF, all distributions are very close except the lowest precision $N=1, B=4$ case. In order to facilitate quantization effects, we plot PDF of the probability error for top-1 class as a difference between floating-point probability and its lower precision counterparts in Figure~\ref{fig:3}(c). Then, we can conclude that $N=3, B=4$ classifier bears slightly lower variance than 8-bit fixed-point model and $N=2, B=4$ classifier obtains slightly higher variance with nearly zero bias in both cases. At the same time, the lowest precision model experiences significant divergence from the baseline without retraining process.

\subsection{Complexity Analysis}

Section~\ref{sec:architecture} concludes that the number of product operations in each convolutional layer can be decreased by a factor of $( \tilde{C} H_f W_f / P )$ with all multiplications replaced by power-of-two shifts. Furthermore, the pipelined ShiftALU is able to compute $(P-1)$ outputs in a single cycle by using $(\lfloor P/2 \rfloor -1)$ shift-by-one and $\lfloor P/2 \rfloor$ sign flipping operations. Hence, ShiftALU further reduces the number of cycles needed to process precomputed tensor by a factor of $P$. Finally, the speed-up in terms of the number of cycles for multiplication operations for the conventional approach and the number of ShiftALU cycles for the proposed approach is equal to $ ( \tilde{C} H_f W_f ) $. 


Table~\ref{tab:2} presents a high-level comparison of how many multiplication cycles (operations) needed for the conventional convolutional layers and how many ShiftALU cycles needed for the three analyzed network models. It can be seen that the SqueezeNet has a 260$\times$ speed-up due to the fact that it employs filter kernel with the dimensions $ H_f, W_f \leq 3 $. At the same time, GoogleNet and ResNet-18 widely use filter kernels with the dimensions equal to 5 and 7. Therefore, the achieved speed-ups are 687$\times$ and 1090$\times$, respectively.

\begin{table}[ht]
	\caption{High-level complexity comparison for popular CNNs}
	\label{tab:2}
	\centering
	\begin{tabular}{lccc}
        \toprule
        Network&\specialcell{Conventional Mult.\\  Million Cycles}&\specialcell{ShiftALU\\Million Cycles} & Speed-Up \\
		\midrule
        SqueezeNet & 410 & 1.58 & 260$\times$ \\
        GoogleNet & 1750 & 2.55 & 687$\times$ \\
        ResNet-18 & 1970 & 1.81 & 1090$\times$ \\
		\bottomrule
	\end{tabular}
\end{table}

\subsection{FPGA Implementation and Power Estimates}

We implemented a computational pipeline of ShiftCNN convolutional layer in RTL starting from the ShiftALU input showed in Figure~\ref{fig:1}(b) to the adder tree output showed in Figure~\ref{fig:2}. Memories for the input, output, weight and bias tensors are intentionally excluded from the design. For comparison, a conventional 8-bit fixed-point (multiplier-based) arithmetic unit was implemented in RTL with the same architectural assumptions. Then, RTL code was compiled for an automotive-grade Xilinx Zynq XA7Z030 device running at 200 MHz clock rate. The input tensor bit-width is 8-bit which is enough for almost lossless feature map representation in dynamic fixed-point format~\citep{gysel}. The resulted outputs (before addition operations) of either ShiftALU or the multiplier-based approach are 16-bits. Bit-width $B$ of the ShiftALU indices are $4$-bits, $N=2$ and parallelization level $\bar{C}=128$.

Table~\ref{tab:3} presents FPGA utilization and power consumption estimates of the implemented computational pipeline in terms of number of lookup tables (LUTs), flip-flops (FFs), digital signal processing blocks (DSPs) and dynamic power measured by Xilinx Vivado tool. Note that the multiplier-based approach is implemented both using DSP blocks and using LUTs only. According to these estimates, ShiftCNN utilizes $2.5\times$ less FPGA resources and consumes $4\times$ less dynamic power. The latter can be explained by low update rate of the precomputed tensor memory. Moreover, our power evaluation shows that 75\% of ShiftCNN dynamic power is dissipated by the adder tree and, hence, the power consumption of product computation is decreased by a factor of 12. In other words, the power consumption of convolutional layer is dominated by addition operations.

\begin{table}[ht]
	\caption{FPGA utilization and power consumption estimates}
	\label{tab:3}
	\centering
	\begin{tabular}{lccccc}
        \toprule
        ALU & LUTs & FFs & DSPs & Power, mW & Relative Power \\
        \midrule
        Adders only & 1644 & 1820 & 0 & 76 & 75\% \\
        ShiftCNN & 4016 & 2219 & 0 & 102 & 100\% \\
        Mult. DSP & 0 & 2048 & 191 & 423 & 415\% \\
        Mult. LUT & 10064 & 4924 & 0 & 391 & 383\% \\
		\bottomrule
	\end{tabular}
\end{table}

\section{Conclusions}
\label{sec:conclusions}

We introduced ShiftCNN, a novel low-precision CNN architecture. Flexible hardware-efficient weight representation allowed to compute convolutions using shift and addition operations only with the option to support binary-quantized and ternary-quantized networks. The algorithm to quantize full-precision floating-point model into ShiftCNN without retraining was presented. In addition, a naturally small codebook of weights formed by the proposed low-precision representation allowed to precompute convolution terms and, hence, to reduce the number of computation operations by two orders of magnitude or more for commonly used CNN models.

ImageNet experiments showed that state-of-the-art CNNs when being converted into ShiftCNN experience less than 1\% drop in accuracy for a reasonable model configuration. Power analysis of the proposed design resulted in a factor of 4 dynamic power consumption reduction of convolutional layers when implemented on a FPGA compared to conventional 8-bit fixed-point architectures. We believe, the proposed architecture is able to surpass existing fixed-point inference accelerators and make a new types of deep learning applications practically feasible for either embedded systems or data center deployments.

\bibliography{paper}

\begin{thebibliography}{20}
\providecommand{\natexlab}[1]{#1}
\providecommand{\url}[1]{\texttt{#1}}
\expandafter\ifx\csname urlstyle\endcsname\relax
  \providecommand{\doi}[1]{doi: #1}\else
  \providecommand{\doi}{doi: \begingroup \urlstyle{rm}\Url}\fi

\bibitem[Bagherinezhad et~al.(2016)Bagherinezhad, Rastegari, and Farhadi]{lcnn}
Hessam Bagherinezhad, Mohammad Rastegari, and Ali Farhadi.
\newblock {LCNN:} lookup-based convolutional neural network.
\newblock \emph{arXiv preprint arXiv:1611.06473}, 2016.

\bibitem[Choi et~al.(2017)Choi, El{-}Khamy, and Lee]{limits}
Yoojin Choi, Mostafa El{-}Khamy, and Jungwon Lee.
\newblock Towards the limit of network quantization.
\newblock \emph{International Conference on Learning Representations (ICLR)},
  Apr 2017.

\bibitem[Courbariaux et~al.(2015)Courbariaux, Bengio, and
  David]{CourbariauxBD14}
Matthieu Courbariaux, Yoshua Bengio, and Jean{-}Pierre David.
\newblock Training deep neural networks with low precision multiplications.
\newblock In \emph{International Conference on Learning Representations
  (ICLR)}, May 2015.

\bibitem[Goodfellow et~al.(2016)Goodfellow, Bengio, and Courville]{goodbook}
Ian Goodfellow, Yoshua Bengio, and Aaron Courville.
\newblock \emph{Deep Learning}.
\newblock MIT Press, 2016.

\bibitem[Gysel et~al.(2016)Gysel, Motamedi, and Ghiasi]{gysel}
Philipp Gysel, Mohammad Motamedi, and Soheil Ghiasi.
\newblock Hardware-oriented approximation of convolutional neural networks.
\newblock \emph{arXiv preprint arXiv:1604.03168}, 2016.

\bibitem[Han et~al.(2016)Han, Mao, and Dally]{dcm}
Song Han, Huizi Mao, and William~J Dally.
\newblock Deep compression: Compressing deep neural networks with pruning,
  trained quantization and {H}uffman coding.
\newblock In \emph{International Conference on Learning Representations
  (ICLR)}, May 2016.

\bibitem[He et~al.(2015)He, Zhang, Ren, and Sun]{he}
Kaiming He, Xiangyu Zhang, Shaoqing Ren, and Jian Sun.
\newblock Deep residual learning for image recognition.
\newblock \emph{arXiv preprint arXiv:1512.03385}, 2015.

\bibitem[Horowitz(2014)]{horowitz}
Mark Horowitz.
\newblock Computing's energy problem (and what we can do about it).
\newblock In \emph{IEEE Interational Solid State Circuits Conference (ISSCC)},
  pages 10--14, Feb 2014.

\bibitem[Hubara et~al.(2016)Hubara, Courbariaux, Soudry, El-Yaniv, and
  Bengio]{courbariauxB16}
Itay Hubara, Matthieu Courbariaux, Daniel Soudry, Ran El-Yaniv, and Yoshua
  Bengio.
\newblock Binarized neural networks.
\newblock In \emph{Advances in Neural Information Processing Systems 29
  (NIPS)}, pages 4107--4115. 2016.

\bibitem[Jia et~al.(2014)Jia, Shelhamer, Donahue, Karayev, Long, Girshick,
  Guadarrama, and Darrell]{caffe}
Yangqing Jia, Evan Shelhamer, Jeff Donahue, Sergey Karayev, Jonathan Long, Ross
  Girshick, Sergio Guadarrama, and Trevor Darrell.
\newblock Caffe: Convolutional architecture for fast feature embedding.
\newblock \emph{arXiv preprint arXiv:1408.5093}, 2014.

\bibitem[Lavin(2015)]{Lavin15b}
Andrew Lavin.
\newblock Fast algorithms for convolutional neural networks.
\newblock \emph{arXiv preprint arXiv:1509.09308}, 2015.

\bibitem[LeCun et~al.(1990)LeCun, Denker, and Solla]{obd}
Yann LeCun, John~S. Denker, and Sara~A. Solla.
\newblock Optimal brain damage.
\newblock In \emph{Advances in Neural Information Processing Systems 2 (NIPS)},
  pages 598--605. 1990.

\bibitem[Li and Liu(2016)]{liL16}
Fengfu Li and Bin Liu.
\newblock Ternary weight networks.
\newblock \emph{arXiv preprint arXiv:1605.04711}, 2016.

\bibitem[Miyashita et~al.(2016)Miyashita, Lee, and Murmann]{MiyashitaLM16}
Daisuke Miyashita, Edward~H. Lee, and Boris Murmann.
\newblock Convolutional neural networks using logarithmic data representation.
\newblock \emph{arXiv preprint arXiv:1603.01025}, 2016.

\bibitem[Rastegari et~al.(2016)Rastegari, Ordonez, Redmon, and
  Farhadi]{RastegariORF16}
Mohammad Rastegari, Vicente Ordonez, Joseph Redmon, and Ali Farhadi.
\newblock {XNOR}-net: Imagenet classification using binary convolutional neural
  networks.
\newblock \emph{arXiv preprint arXiv:1603.05279}, 2016.

\bibitem[Redmon et~al.(2016)Redmon, Divvala, Girshick, and Farhadi]{yolo}
Joseph Redmon, Santosh Divvala, Ross Girshick, and Ali Farhadi.
\newblock You only look once: Unified, real-time object detection.
\newblock In \emph{The IEEE Conference on Computer Vision and Pattern
  Recognition (CVPR)}, June 2016.

\bibitem[Russakovsky et~al.(2015)Russakovsky, Deng, Su, Krause, Satheesh, Ma,
  Huang, Karpathy, Khosla, Bernstein, Berg, and Fei-Fei]{imagenet}
Olga Russakovsky, Jia Deng, Hao Su, Jonathan Krause, Sanjeev Satheesh, Sean Ma,
  Zhiheng Huang, Andrej Karpathy, Aditya Khosla, Michael Bernstein,
  Alexander~C. Berg, and Li~Fei-Fei.
\newblock {ImageNet Large Scale Visual Recognition Challenge}.
\newblock \emph{International Journal of Computer Vision (IJCV)}, 115\penalty0
  (3):\penalty0 211--252, 2015.

\bibitem[Shelhamer et~al.(2016)Shelhamer, Long, and Darrell]{long}
Evan Shelhamer, Jonathan Long, and Trevor Darrell.
\newblock Fully convolutional networks for semantic segmentation.
\newblock \emph{arXiv preprint arXiv:1605.06211}, 2016.

\bibitem[Sze et~al.(2017)Sze, Chen, Yang, and Emer]{survey}
Vivienne Sze, Yu-Hsin Chen, Tien-Ju Yang, and Joel Emer.
\newblock Efficient processing of deep neural networks: A tutorial and survey.
\newblock \emph{arXiv preprint arXiv:1703.09039}, 2017.

\bibitem[Zhou et~al.(2017)Zhou, Yao, Guo, Xu, and Chen]{zhou2017}
Aojun Zhou, Anbang Yao, Yiwen Guo, Lin Xu, and Yurong Chen.
\newblock Incremental network quantization: Towards lossless {CNN}s with
  low-precision weights.
\newblock In \emph{International Conference on Learning Representations
  (ICLR)}, Apr 2017.

\end{thebibliography}

\end{document}